\documentclass[11pt]{article}
\usepackage{kotex}
\usepackage[final]{acl}
\usepackage{times}
\usepackage{latexsym}
\usepackage[T1]{fontenc}
\usepackage[utf8]{inputenc}
\usepackage{microtype}
\usepackage{inconsolata}
\usepackage{graphicx}
\usepackage{booktabs}
\usepackage{multirow}
\usepackage{amsmath}
\usepackage{xcolor}
\usepackage{url}
\usepackage{tabularx}
\usepackage{tikz}
\usetikzlibrary{arrows.meta,positioning}

\graphicspath{{figure/}}

\newcommand{\cypher}[1]{\texttt{#1}}

\title{KG2Cypher: Data-Centric Pipeline for Building Enterprise Text-to-Cypher Systems}

\author{
  Minjun Choi$^{1,\dagger}$ \quad 
  Yerin Kim$^{2,\dagger}$ \quad 
  Junghyuk Seo$^2$ \quad 
  Sujin Mo$^2$ \quad 
  Hyemin Lee$^2$ \\ 
  \textbf{Youngjoong Ko}$^1$\thanks{Corresponding author.} \\[0.15cm]
  $^1$Sungkyunkwan University \quad $^2$NAVER \\
   \texttt{\{alswns078, lovekyll0\}@gmail.com, yjko@skku.edu} \\
   \texttt{\{junghyuk.seo, sujin.mo, hmin.lee\}@navercorp.com}
}

\begin{document}
\maketitle

\bgroup
\renewcommand{\thefootnote}{\ensuremath{\dagger}}
\footnotetext{ This work was done while Minjun Choi and Yerin Kim were Research Interns at NAVER.}
\egroup

\begin{abstract}
Enterprise Knowledge Graphs (KGs) are increasingly used for internal search, analytics, and question answering, but building natural-language interfaces for private enterprise graphs remains costly. We present KG2Cypher, a data-centric pipeline for building enterprise text-to-Cypher systems from existing KGs. KG2Cypher first constructs an executable Cypher query from observed graph facts and then uses LLMs to generate its associated natural-language question. The resulting Text-Cypher pairs are validated with an LLM judge and human validation, and are converted into candidate-aware SFT data. The trained generator is served with class-conditioned schema prompting, entity retrieval, and LoRA-based inference. We evaluate KG2Cypher in Korean enterprise settings, where short search-style queries and schema paraphrases make language grounding difficult. LoRA SFT improves execution-result F1 from 0.806 to 0.950 on broadcast-program queries and from 0.70 to 0.92 on company queries. In an 11-class setting, KG2Cypher achieves 95.2\% exact match, 99.9\% execution rate, and 0.964 execution-result F1.
\end{abstract}

\section{Introduction}

Enterprise Knowledge Graphs (KGs) store structured business knowledge for internal search, analytics, question answering, and so on. For example, a media KG may connect a program to its broadcaster, genre, cast, and episode count and a company KG may connect an organization to its industry, founders, listing exchange, and financial attributes. These graphs are very useful, but most users do not know which node types and relation types exist, how to write Cypher queries (hereafter, Cypher), or which internal IDs identify the entities in the graph. This creates a need for a natural-language interface that lets users outside expert teams use enterprise KGs as well.

\begin{table}[t]
\centering
\scriptsize
\setlength{\tabcolsep}{3pt}
\renewcommand{\arraystretch}{1.08}
\begin{tabularx}{\columnwidth}{@{}p{0.26\columnwidth}X@{}}
\toprule
\multicolumn{2}{@{}l}{\textbf{Broadcast query}} \\
\midrule
\textsc{Text (Input)} &
고윤정이 출연한 24부작 JTBC 방송 프로그램
\newline
(24-episode JTBC broadcast programs featuring Go Youn-jung) \\
\textsc{Required Grounding} &
Go Youn-jung URI + JTBC URI + episode-count literal (= 24) \\
\textsc{Cypher (Output)} &
\texttt{MATCH ... cast\_member ... broadcast\_by ... number\_of\_episodes = 24} \\
\midrule
\multicolumn{2}{@{}l}{\textbf{Company query}} \\
\midrule
\textsc{Text (Input)} &
초봉 4천만원 이상의 코스피 상장 중견기업
\newline
(mid-sized companies listed on KOSPI with starting salary of at least 40M KRW) \\
\textsc{Required Grounding} &
salary literal ($\geq$ 40000000) + KOSPI URI + legal-form URI \\
\textsc{Cypher (Output)} &
\texttt{MATCH ... first\_salary ... stock\_exchange ... legal\_form} \\
\bottomrule
\end{tabularx}
\caption{Illustrative Korean text-to-Cypher examples. English translations are shown in parentheses. Cypher outputs are anonymized because the underlying enterprise KG and entity identifiers are private.}
\label{tab:intro-examples}
\end{table}

Text-to-Cypher transformation is a constrained structured-query generation task. A generated query has to choose valid schema relations, construct graph patterns, bind entity URIs, use the correct literal sub-fields, and execute against a live graph database. Table~\ref{tab:intro-examples} illustrates these requirements with broadcast and company queries from our Korean enterprise settings. These examples show how user expressions must be mapped to entity URIs, schema relations, literal conditions, and executable Cypher. Users may write short search-style phrases, omit arguments, vary spacing, mix Korean text with transliterated or foreign names, and use Korean paraphrases that do not match schema relation names.

These constraints make data construction a central challenge. The most straightforward solution is manual annotation, but this lacks scalability in an enterprise knowledge graph (KG) environment because each domain has its own node types, relation names, entity identifiers, and literal conventions. For instance, a dataset for broadcast programs does not cover companies, sports teams, or festivals. As a result, the manual construction of natural-language and Cypher pairs requires a separate annotation task for each new domain. Furthermore, in-context learning with a strong LLM can be another solution. However, a prompt-only gpt-oss-120B model often produced syntactically executable Cypher, but it still returned wrong graph results because it selected the wrong relation, hallucinated an entity identifier, or used the wrong literal format in our experiments.

Our key idea is simple; an enterprise that already has a KG should be able to reuse the KG itself as a supervision source. We present KG2Cypher, a data-centric industry pipeline that implements this idea for building enterprise text-to-Cypher systems. In the data construction stage, KG2Cypher samples relation patterns that appear in the graph, executes those patterns to obtain real subgraphs, and builds executable Cypher using the returned entities and literals. LLMs are then used only for language-side operations including paraphrasing, compressed query generation, and quality judging. This automates most of the symbol manipulation, and it can reduce the need to manually construct Text-Cypher pairs from scratch. As a result, human efforts are focused more on verification and revision than on initial data creation.

During the training and serving stages, KG2Cypher converts verified pairs into candidate-aware SFT examples. The prompt contains the question, candidate relations from the class schema, and entity candidates with retrieval distractors, so the model learns to select the needed relations and URIs. A LoRA adapter is then trained and served with the same prompt structure in a production-oriented inference pipeline.

We evaluate KG2Cypher on proprietary Korean enterprise KG domains including broadcast and company settings. Although our experiments use Korean queries, the pipeline is not designed only for Korean because relation sampling, subgraph fetching, and canonical Cypher construction operate only on graph structure and graph values. To apply KG2Cypher to another enterprise language, the language-facing components need adaptation including question diversification prompts, judge prompts, and the domain classifier.
Overall, the experiments show that execution validity alone is not sufficient, because prompt-only models can run but return wrong results. KG-grounded SFT improves enterprise-specific grounding, and class-conditioned schema prompting avoids relation-first retrieval in our service setting.

\begin{figure*}[t]
\centering
\includegraphics[width=0.98\textwidth]{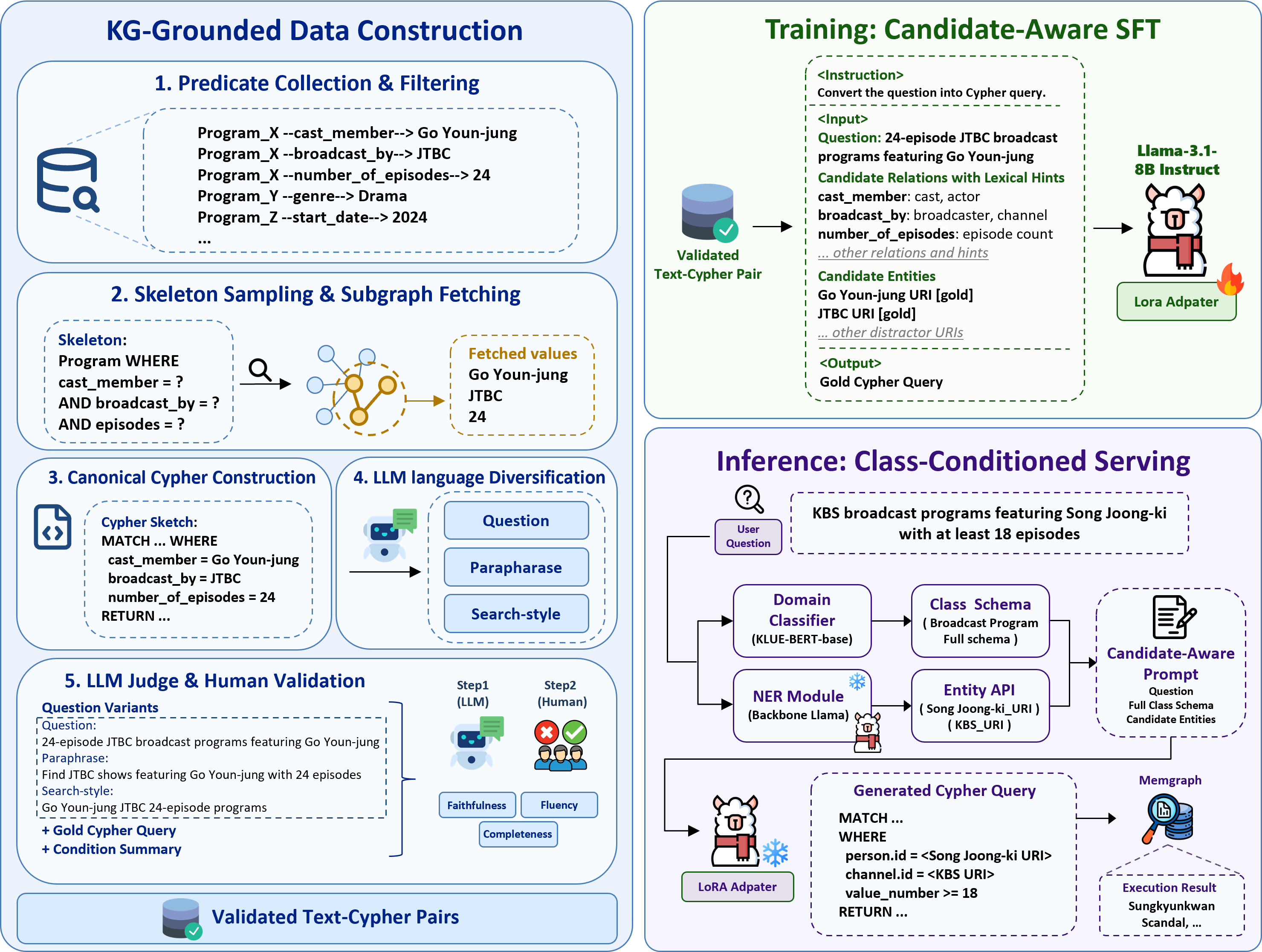}
\caption{Overview of KG2Cypher. Left: KG-grounded data construction from graph facts to validated Text-Cypher pairs. Right: candidate-aware SFT and class-conditioned serving. Examples are translated, and Cypher is anonymized.}
\label{fig:kg2cypher-overview}
\end{figure*}

\section{Related Work}

\paragraph{Structured query generation.}
Natural-language interfaces to databases have long been studied as structured query generation. Text-to-SQL benchmarks such as WikiSQL and Spider define tasks that map user questions to executable SQL queries \citep{zhong2017seq2sql,yu2018spider}. Later methods study schema linking, constrained decoding, and LLM prompting for more reliable query generation \citep{wang2020ratsql,scholak2021picard,gao2024texttosql}. KGQA datasets such as WebQuestionsSP, LC-QuAD, and GrailQA also map natural-language questions to logical forms or graph queries \citep{yih2016value,dubey2019lcquad,gu2021grailqa}. These works establish evaluation practices for executable structured queries. KG2Cypher follows this tradition, but the target is Cypher over a property graph. The system must also bind private entity URIs and use enterprise-specific relation names.

\paragraph{Text-to-Cypher and enterprise graph settings.}
Cypher is a property-graph query language for expressive graph pattern matching in industrial graph databases \citep{francis2018cypher}. Recent Text-to-Cypher work addresses the lack of public data and evaluation resources. The Neo4j Text2Cypher dataset combines public examples into a large benchmark \citep{ozsoy2025text2cypher}. Auto-Cypher/SynthCypher uses LLM-supervised generation and verification for synthetic Cypher data \citep{tiwari2025autocypher}. Mind the Query emphasizes execution-grounded benchmarking with graph databases and validation checks \citep{chauhan2025mind}. Recent multilingual Text-to-Cypher work also reports performance gaps across languages \citep{ozsoy2025multilingual}. These studies make Text-to-Cypher more measurable on public resources. KG2Cypher addresses a different industry problem. It constructs data, trains a model, and supports deployment for private enterprise KGs whose data, identifiers, schemas, and retrieval APIs cannot be released.

\section{Methodology}

\subsection{Task Formulation}

Given a natural-language question $q$, relation candidates $R$, and entity candidates $E$, the generator model $f_\theta$ produces an executable Cypher query $y$:
\begin{equation}
    y = f_\theta(q, R, E).
\end{equation}
This formulation aligns supervised fine-tuning (SFT) with deployment. The generator must select valid schema elements from retrieved candidates rather than generate them from scratch. Each relation candidate $r \in R$ contains subject and object classes, a predicate identifier, and linguistic hints. Each entity candidate $e \in E$ contains an internal URI, display name, and class label.

\subsection{System Overview}

Figure~\ref{fig:kg2cypher-overview} outlines the full workflow of KG2Cypher: KG-grounded data construction, candidate-aware SFT, and class-conditioned serving.
The key design choice is to separate symbolic query construction from language generation. KG2Cypher builds Cypher targets from graph values with deterministic code and uses LLMs for paraphrasing and validation. This reduces failures such as nonexistent relations, hallucinated entity identifiers, and literal conditions that do not execute.

\subsection{Predicate Collection and Filtering}

Instead of relying on static schema specifications, the pipeline inspects graph instances to collect Subject-Predicate-Object (SPO) patterns that connect subject and object nodes. For the broadcast-program class, this step identifies active relations with predicate identifiers such as ``broadcast\_by'', ``genre'', and ``number\_of\_episodes'' and it also records whether each object is an entity or a literal. This ensures that subsequent queries are based on the observed graph facts.
Rule-based filtering removes metadata and non-searchable attributes such as geocoordinates, media URLs, social-media IDs, and system fields. Objects with the same relation identifiers are merged if their query semantics are equivalent.

\subsection{Skeleton Sampling and Subgraph Fetching}

The filtered relations are combined into multi-condition query skeletons with the bucket distributions 40/30/20/10 for one-, two-, three-, and four-relation structures, respectively. Skeletons are discarded if a domain lacks sufficient relations, and a limit on the number of attempts prevents redundant sampling. Each skeleton is validated against the Memgraph graph database via a \cypher{LIMIT 1} query. For each valid skeleton, the pipeline samples up to 50 matching subgraphs. This limit prevents high-frequency graph patterns from biasing the dataset and collects real entity URIs, literals, and relation attributes for grounded data generation.

\subsection{Canonical Cypher Construction}

KG2Cypher deterministically constructs a canonical target $C_{\text{gold}}$ for each subgraph. Entity nodes are bound by unique graph identifiers in the \cypher{WHERE} clause, and literals are mapped to schema attributes with valid comparison operators $\theta \in \{=, >, <, \ge, \le\}$.
This stage also creates an analyzed intermediate form $\text{NL}_{\text{analyzed}}$ and a template-derived naive statement $\text{NL}_{\text{naive}}$ alongside $C_{\text{gold}}$. These synchronized views expose the same query semantics and keep the next LLM rewriting step anchored to verified graph structure and literal constraints. Appendix~\ref{sec:data-construction-example} gives a concrete example of these representations.

\subsection{LLM-Based Language Diversification}

This stage uses the synchronized representations ($C_{\text{gold}}$, $\text{NL}_{\text{analyzed}}$, $\text{NL}_{\text{naive}}$), ontology constraints, and target-language synonym maps as input to gpt-oss-120B. KG2Cypher keeps the symbolic Cypher target fixed and uses the LLM only to rewrite the language side. This design reduces unsupported graph structures and hallucinated literal constraints.

This language-side expansion is related to self-instruction \citep{wang2023selfinstruct}, but KG2Cypher fixes the symbolic Cypher target before rewriting.
The LLM generates three types of questions: a term-preserving question, five paraphrases, and compressed search-style queries for shallow skeletons ($\le 3$ joins). For numeric and date relations, deterministic checks verify that units and comparison words match the Cypher condition, such as mapping ``at least'' to $\ge$.

\subsection{LLM Judge and Human Validation}

To detect semantic drift, KG2Cypher uses gpt-oss-120B to score each instance on a 0/1/2 scale across faithfulness to $C_{\text{gold}}$, target-language fluency, and completeness of schema constraints. The scale is a simple ordinal rubric used to align LLM scores with human validation labels. Instances that pass all dimensions receive a \texttt{pass} status. Imperfect rows are marked as \texttt{needs\_review} and routed to a human validation interface for verification (\texttt{keep}) or correction (\texttt{edit}).

The judge is calibrated based on human-assigned scores and brief comments that explain the reason for score deductions from 200 sampled instances. KG2Cypher automatically revises the scoring guide prompt with gpt-oss-120B using these comments. Because the validated synthetic data has high-score skew and low variance, we use Mean Absolute Error (MAE), adjacent agreement, and deduction catch rate instead of variance-dependent metrics, in line with concerns from LLM-as-a-judge work \citep{zheng2023judging}. Appendix~\ref{sec:appendix-failure-scenarios} shows why no single metric is sufficient.

\subsection{Candidate-Aware SFT Construction}

After the construction and validation stages, KG2Cypher converts validated Text-Cypher pairs into instruction-following SFT examples. Each input contains the question, candidate relations, and candidate entities, and the output is the gold Cypher. Candidate relations include subject and object classes, predicate identifiers, and linguistic hints. Candidate entities include URI, name, class, and retrieval distractors from the inference-time entity API. This matches inference-time prompts, exposes the model to retrieval noise, and forces it to select the relations and entity URIs needed for the question. Appendix~\ref{sec:sft-example} gives a full anonymized example.

\subsection{Class-Conditioned Schema Prompting}

Prior KBQA systems often retrieve candidate relations before logical-form generation. For example, SG-KBQA ranks question-relation pairs with a BERT-based cross-encoder \citep{gao2025sgkbqa}. This relation-first design is effective in benchmarks, but it is difficult to use as-is in a low-latency enterprise service.

In practice, relation retrieval is not reliable enough in our low-latency enterprise service setting. If the gold relation is absent from a prompt, the generator usually cannot recover. In contrast, entity retrieval is reliable enough to provide URI candidates. Thus KG2Cypher uses class-conditioned schema prompting. A domain classifier first selects the target graph class and then the prompt includes the full relation schema for that class, while entity candidates still come from the entity API.

The KLUE-BERT-base model \citep{park2021klue} is trained for 11-way single-label classification as a domain classifier. This design shifts retrieval from relation-level ranking to class routing and entity URI binding, and it keeps the prompt bounded by the predicted class schema.

\subsection{Training and Serving}

The generator is based on the Llama-3.1-8B-Instruct model \citep{grattafiori2024llama}. We adapt it with LoRA \citep{hu2022lora}. The final adapter trains approximately 41M parameters, about 0.52\% of the base model. Prompt tokens are masked, and loss is computed only on the assistant Cypher response. Appendix~\ref{sec:training-details} provides the full training configuration.

At inference time, the system classifies the domain, loads the class schema, performs NER, retrieves entity candidates, builds the prompt, and generates Cypher through vLLM \citep{kwon2023vllm}. The backbone stays loaded in vLLM; NER uses it without LoRA, and Cypher generation attaches the LoRA adapter at request time to avoid separate large-model endpoints. For entity retrieval, KG2Cypher merges two API result sets, one from the full question with detected entities and the other from detected entities alone.

\section{Experiments}

\subsection{Data and Metrics}

We evaluate KG2Cypher with internal enterprise KG domains. The final class-conditioned setting uses 11 graph classes with 20,745/2,590/2,621 train/dev/test examples; Appendix~\ref{sec:class-labels} lists the classes. We also report broadcast and company diagnostics for prompting, retrieval, SFT, and transfer.
We report EM, execution rate, and execution-result F1. EM is exact string match; execution rate checks whether Cypher runs on Memgraph; and execution-result F1 compares predicted and gold answer sets. We use F1 as the main user-facing metric because executable Cypher can return wrong results. Appendix~\ref{sec:metric-example} gives an anonymized scoring example.

\subsection{LLM Judge Calibration}

Before using the LLM judge to validate generated Text-Cypher pairs, we calibrate its scoring prompt on 200 human-annotated samples. The judge scores faithfulness, fluency, and completeness, and calibration checks agreement with human labels. We mainly track MAE, where lower values mean closer agreement with human scores. The initial prompt passes fluency and completeness but fails faithfulness with MAE 0.314, so it is not reliable enough for automatic quality gating. Appendix~\ref{sec:appendix-failure-scenarios} reports the full initial scores and explains the additional agreement checks.

\begin{figure}[t]
\centering
\includegraphics[width=\columnwidth]{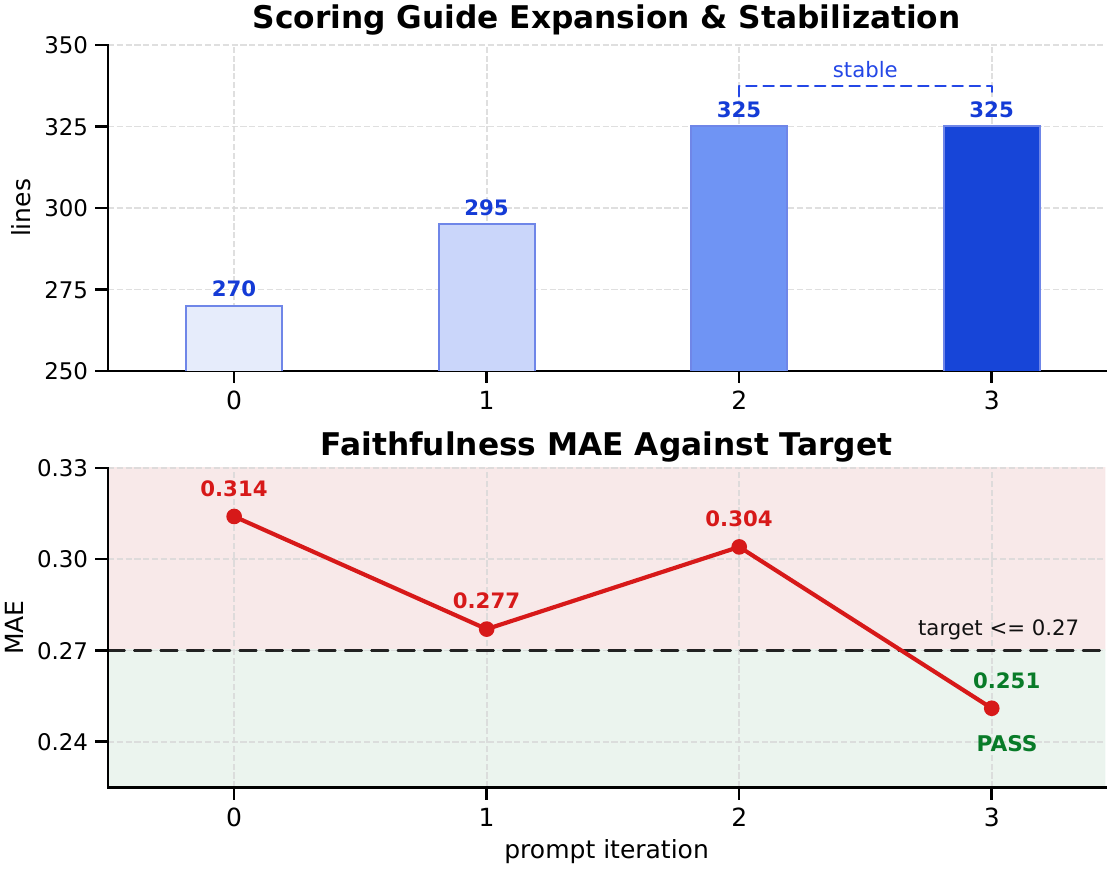}
\caption{LLM judge calibration. Human comments guide automatic scoring-guide revision, and faithfulness MAE falls below the 0.27 target. Lower MAE is better.}\label{fig:judge-calibration-v3}
\end{figure}

KG2Cypher then automatically revises the scoring guide prompt with gpt-oss-120B using the human disagreement comments. As shown in Figure~\ref{fig:judge-calibration-v3}, the guide grows from 270 to 325 lines and then stabilizes. Faithfulness MAE decreases from 0.314 to 0.251 after three iterations, passing the 0.27 target. The calibrated judge accepts high-confidence pairs and routes lower-confidence pairs to human validation.

\subsection{Prompt-Only Diagnostics}

We first test whether a strong prompt-only LLM can generate enterprise Cypher without task-specific training. We use gpt-oss-120B and compare four settings: rules only, rules with five few-shot examples, rules with few-shot examples plus the full broadcast schema, and an oracle prompt with gold relations and gold entities.

\begin{table}[t]
\centering
\footnotesize
\setlength{\tabcolsep}{3pt}
\begin{tabular}{lccc}
\toprule
Prompt setting & EM & Exec. & F1 \\
\midrule
Rules only & 0.0 & 100.0 & 0.0024 \\
Rules + few-shot & 0.5 & 100.0 & 0.0049 \\
Rules + few-shot + schema & 4.2 & 99.7 & 0.0655 \\
Oracle gold candidates & 47.7 & 99.7 & 0.7139 \\
\bottomrule
\end{tabular}
\caption{Broadcast-domain prompt-only diagnostics. EM and execution are percentages; oracle uses gold relation/entity candidates.}
\label{tab:prompt-diagnostics-v4}
\end{table}

Table~\ref{tab:prompt-diagnostics-v4} shows that rules and few-shot examples are not enough. The prompts produce executable Cypher, but EM and execution-result F1 remain near zero, which indicates that the model mostly learns runnable query forms rather than correct graph grounding. Adding the full broadcast schema helps only slightly because relation selection remains unresolved. Oracle gold relation/entity candidates sharply improve performance, which shows that candidate selection is critical. Even with these oracle candidates, prompt-only generation reaches only 47.7\% EM and 0.7139 F1.

We observe three recurring failure modes: hallucinated private entity URIs, plausible but wrong relations, and incorrect literal sub-fields such as \cypher{value\_number} and \cypher{value\_date.month}. These failures motivate retrieval-aware prompting and SFT. Appendices~\ref{sec:prompt-failure} and~\ref{sec:oracle-complexity} give examples and query-complexity results.

\subsection{Retrieval Diagnostics}

Because Table~\ref{tab:prompt-diagnostics-v4} shows that candidate selection is critical, we next evaluate whether available service APIs can retrieve those candidates. Table~\ref{tab:retrieval-v4} reports candidate retrieval diagnostics. Entity retrieval is evaluated on searchable entity mentions because literal-only conditions do not require lookup, and it reaches 95.0\% recall@20. In contrast, the deployed relation API reaches 40.6\% recall@20; query rewriting with five variants raises this to 49.8\%, still too low when generation requires the gold relation.

\begin{table}[t]
\centering
\footnotesize
\setlength{\tabcolsep}{3pt}
\begin{tabular}{llcc}
\toprule
Component & Setting & $k$ & Recall \\
\midrule
Entity & API retrieval & 1 & 67.5\% \\
Entity & API retrieval & 5 & 87.5\% \\
Entity & API retrieval & 20 & 95.0\% \\
\midrule
Relation & deployed API & 1 & 18.8\% \\
Relation & deployed API & 5 & 36.0\% \\
Relation & deployed API & 10 & 39.8\% \\
Relation & deployed API & 20 & 40.6\% \\
Relation & API + query rewriting & 20 & 49.8\% \\
\bottomrule
\end{tabular}
\caption{Retrieval diagnostics. Entity retrieval is reliable, but relation retrieval is the bottleneck.}
\label{tab:retrieval-v4}
\end{table}

This result explains why KG2Cypher does not rely on relation-first retrieval at inference time. In knowledge-base QA, relation-first methods can use heavy cross-encoders to rank question-relation pairs, but our enterprise service setting requires low-latency retrieval. Therefore, KG2Cypher predicts the target class and provides the full relation schema for that class.

\subsection{LoRA SFT and Class-Conditioned Results}
Following the prompt-only and retrieval diagnostics, Table~\ref{tab:sft-v4} reports the main generation results: LoRA SFT on broadcast and company, and the final 11-class class-conditioned setting. Prompt-only baselines receive few-shot examples and gold relation/entity candidates. For company, we also test out-of-domain transfer by applying a broadcast-trained adapter to company queries.

\begin{table}[t]
\centering
\footnotesize
\setlength{\tabcolsep}{2.5pt}
\begin{tabularx}{\columnwidth}{@{}lXccc@{}}
\toprule
Domain & System & EM & Exec. & F1 \\
\midrule
Broadcast & Prompt + gold & 51.5 & 51.5 & 0.806 \\
Broadcast & LoRA SFT & \textbf{86.5} & \textbf{99.5} & \textbf{0.950} \\
\midrule
Company & Prompt + gold & 54.7 & 78.0 & 0.70 \\
Company & LoRA SFT & \textbf{86.9} & \textbf{100.0} & \textbf{0.92} \\
Company & LoRA OOD & 65.3 & 98.8 & 0.79 \\
\midrule
11-class & LoRA + class schema & \textbf{95.2} & \textbf{99.9} & \textbf{0.964} \\
\bottomrule
\end{tabularx}
\caption{Main generation results. EM and execution are percentages; Prompt + gold uses gold relation/entity candidates.}
\label{tab:sft-v4}
\end{table}

LoRA SFT substantially improves execution-result F1 in both broadcast and company domains. The broadcast-trained adapter (LoRA OOD) outperforms prompt-only generation on company queries, but it remains below company in-domain SFT. This shows that Cypher conventions transfer across domains, but class-specific grounding differs enough to motivate training on all target classes and class-conditioned serving. Appendix~\ref{sec:sft-diagnostics} reports a controlled broadcast ablation with distractor relations.

For the final KG2Cypher setting, a KLUE-BERT-base classifier chooses one of 11 graph classes, and the generator receives the schema of that class. The classifier reaches 99.66\% accuracy. The final row of Table~\ref{tab:sft-v4} shows 95.2\% EM, 99.9\% execution rate, and 0.964 execution-result F1. This result supports the class-conditioned design; it avoids low-recall relation retrieval before generation while keeping the prompt bounded by the predicted class. Appendix~\ref{sec:class-labels} lists the 11 graph classes.

Overall, the experiments support three conclusions. First, execution validity alone is not sufficient because prompt-only models can produce executable Cypher that returns wrong graph results. Second, KG-grounded SFT teaches enterprise-specific conventions, including URI binding, relation selection, and literal sub-field use. Third, class-conditioned schema prompting is more practical than relation-first retrieval in our service setting.

\section{Conclusion}

We presented KG2Cypher, a data-centric pipeline that reuses enterprise KGs for Text-to-Cypher data construction, model training, and serving. Its main advantages are that it turns observed graph facts into executable Cypher targets and uses LLMs only for language generation and validation, which reduces manual pair authoring for private enterprise schemas. KG2Cypher then trains a candidate-aware LoRA generator for production-style prompts. Experiments show strong gains over prompt-only generation and demonstrate that class-conditioned schema prompting avoids a major relation-retrieval bottleneck.

\section{Limitations}

The KG, entity APIs, generated data, and model checkpoints are proprietary, so we report aggregate statistics and anonymized examples rather than released artifacts. The pipeline also inherits KG coverage limits: missing relations, entities, or values cannot produce supervision. Although the symbolic stages are language-agnostic, our current instantiation includes language-specific prompts, judge calibration data, and classification modules because the enterprise data are Korean. Applying the system to other languages would require adapting these language-facing components.

Deployment still depends on domain classification, NER, entity retrieval, and entity disambiguation. Outside entities and homonymous entities remain challenging cases. A practical extension is to treat frequent outside values as name-value fields, but scaling this solution across all classes requires further engineering. We evaluate the generator with Llama-3.1-8B-Instruct, and further experiments are needed to assess transfer to other base models.

\section*{Ethical Considerations}

The system is intended for enterprise KG search and analytics by authorized users. Because internal KGs may contain proprietary or sensitive business information, generated data, execution logs, and model outputs must follow organizational access-control and data-governance policies. Human validation is used to reduce semantic drift in generated questions. Examples in this paper are anonymized to avoid exposing private identifiers, and Korean examples are translated for readability.

\bibliography{custom}

\appendix

\section{LLM Judge Metric Validation}
\label{sec:appendix-failure-scenarios}

We use synthetic failure simulations to check why the calibration gate uses MAE, adjacent agreement, and catch rate together. Each simulation contains 200 virtual samples and represents a common judge failure pattern.

Single metrics miss different errors. In a \textit{Systemic Bias Scenario}, the judge assigns every score one point too low. Adjacent agreement can still be 100\%, but MAE detects the bias. In a \textit{Random-Guessing Scenario}, the judge may catch some deducted cases by chance, but MAE and adjacent agreement fail. These examples show why KG2Cypher requires the judge to satisfy all three criteria before the prompt is accepted for data validation.

Table~\ref{tab:judge-v3} reports the initial judge scores before calibration.

\begin{table}[h]
\centering
\small
\begin{tabular}{lcccc}
\toprule
Judge setting & MAE & Adj. & Catch & Pass \\
\midrule
Faith., base     & 0.314 & 1.000 & 0.60 & No \\
Fluency, base    & 0.216 & 0.995 & 0.54 & Yes \\
Complete., base  & 0.000 & 1.000 & --   & Yes \\
\bottomrule
\end{tabular}
\caption{Initial baseline evaluator performance on 200 human-annotated samples before prompt calibration.}
\label{tab:judge-v3}
\end{table}

\section{Evaluation Metric Example}
\label{sec:metric-example}

Execution-result F1 compares the answer set returned by the predicted Cypher with the answer set returned by the gold Cypher. For example, assume the gold query returns four programs:
\[
\{\text{Program A}, \text{Program B}, \text{Program C}, \text{Program D}\}
\]
and the predicted query returns three programs:
\[
\{\text{Program A}, \text{Program B}, \text{Program E}\}.
\]
The true positives are Program A and Program B. Precision is $2/3$, recall is $2/4$, and F1 is the harmonic mean of precision and recall.

\section{Prompting Failure Modes}
\label{sec:prompt-failure}

The prompt-only diagnostics reveal three common failure modes. First, the model may hallucinate private entity identifiers because the identifiers are not inferable from surface text alone. Second, the model may choose a plausible but wrong relation even when the schema is present. For example, a query about an ordered broadcast episode can be confused with an episode-count relation. Third, literal nodes require schema-specific sub-fields. Date, number, price, and time expressions must map to fields such as \cypher{value\_date.month}, \cypher{value\_number}, or a value-unit pair. These cases motivate candidate-aware SFT.

\section{Class Labels}
\label{sec:class-labels}

The 11-class class-conditioned setting uses the following graph classes: broadcast program, company, person, performance, music song, automobile model, movie, sports team, organization, country, and festival. These labels are the output space of the domain classifier. At inference time, the predicted class determines which full relation schema is inserted into the generator prompt.

\section{Oracle Prompting by Query Complexity}
\label{sec:oracle-complexity}

Table~\ref{tab:complexity-v4} breaks down oracle prompt-only performance by the number of relations in the target query.

\begin{table}[h]
\centering
\small
\begin{tabular}{lcc}
\toprule
Relation count & Failure rate & Avg. F1 \\
\midrule
1 & 19\% & 0.8221 \\
2 & 39\% & 0.6544 \\
3 & 42\% & 0.6818 \\
4 & 73\% & 0.2667 \\
\bottomrule
\end{tabular}
\caption{Oracle prompt-only performance by query complexity. Failure increases as more relations must be composed.}
\label{tab:complexity-v4}
\end{table}

\section{Candidate-Aware SFT Diagnostics}
\label{sec:sft-diagnostics}

Before the final class-conditioned setting, we analyze candidate-aware SFT under controlled broadcast-domain conditions. We compare two schema formats. The CLS format includes the target class of each relation, such as \cypher{[D] BROADCAST\_PROGRAM [N] narrator [R] PERSON}. The No-CLS format replaces the target class with a coarse type such as \cypher{ENTITY} or \cypher{LITERAL}. We also compare two candidate conditions. The oracle condition provides only the gold relations. The noise condition adds three negative relation candidates for each gold relation.

\begin{table}[h]
\centering
\small
\setlength{\tabcolsep}{3pt}
\begin{tabular}{llccc}
\toprule
Candidates & Schema & EM & Exec. & F1 \\
\midrule
Oracle & CLS & 89.8 & 99.8 & 0.9499 \\
Oracle & No-CLS & 86.5 & 99.5 & 0.9509 \\
Noise & CLS & 86.0 & 99.8 & 0.9138 \\
Noise & No-CLS & 83.5 & 99.8 & 0.9082 \\
\bottomrule
\end{tabular}
\caption{Broadcast-domain SFT diagnostics. EM and execution are percentages. Noise candidates reduce F1, which supports training and evaluation with retrieval-like distractors.}
\label{tab:sft-diagnostics-v4}
\end{table}

Table~\ref{tab:sft-diagnostics-v4} shows that SFT remains robust with noisy relation candidates, but noise clearly reduces execution-result F1. This supports the final candidate-aware training setup, where the model must choose from realistic relation and entity candidates. Table~\ref{tab:sft-example} gives an anonymized SFT instance with candidate relations, entity candidates, and distractors.

\section{Training Details}
\label{sec:training-details}

Table~\ref{tab:training-details} lists the LoRA training configuration used for the generator. The setup trains only adapter parameters and masks prompt tokens, so the loss is applied to the assistant-side Cypher response rather than to the input context.

\begin{table}[h]
\centering
\small
\begin{tabular}{ll}
\toprule
Hyperparameter & Value \\
\midrule
Base model & Llama-3.1-8B-Instruct \\
Trainable parameters & 41M / 8B (0.52\%) \\
Maximum sequence length & 4096 \\
LoRA rank & 16 \\
LoRA alpha & 32 \\
LoRA dropout & 0.05 \\
Learning rate & $2\times10^{-4}$ \\
Micro-batch size & 1 \\
Gradient accumulation & 8 \\
Devices & 8 NVIDIA A100 GPUs \\
Precision & bf16 mixed \\
\bottomrule
\end{tabular}
\caption{LoRA training configuration.}
\label{tab:training-details}
\end{table}

\section{Data Construction Example}
\label{sec:data-construction-example}

Table~\ref{tab:data-construction-example} illustrates how one grounded graph record is transformed before SFT. The canonical Cypher $C_{\text{gold}}$ is built first, and the language-side forms $\text{NL}_{\text{analyzed}}$ and $\text{NL}_{\text{naive}}$ are derived from the same verified conditions. The LLM then rewrites only the question text, while the gold Cypher remains fixed.

\begin{table*}[p]
\centering
\small
\setlength{\tabcolsep}{4pt}
\renewcommand{\arraystretch}{1.12}
\begin{tabularx}{0.98\linewidth}{@{}lX@{}}
\toprule
Stage & Example \\
\midrule
Fetched values &
Person-Y, Channel-X, and episode count 24 from a broadcast-program subgraph. \\[2pt]

Gold Cypher ($C_{\text{gold}}$) &
\begin{minipage}[t]{\linewidth}
\ttfamily\small
MATCH ... cast\_member ... broadcast\_by ... number\_of\_episodes ...\\
WHERE person.id = <PERSON\_URI\_001>\\
\phantom{WHERE }AND channel.id = <CHANNEL\_URI\_001>\\
\phantom{WHERE }AND value\_number = 24\\
RETURN ...
\end{minipage} \\
Analyzed form ($\text{NL}_{\text{analyzed}}$) &
Target: broadcast program. Conditions: cast member is Person-Y, broadcaster is Channel-X, and episode count is 24. \\
Naive statement ($\text{NL}_{\text{naive}}$) &
Find broadcast programs whose cast member is Person-Y, broadcaster is Channel-X, and episode count is 24. \\[2pt]
LLM question outputs &
\textbf{Question}: 24-episode Channel-X broadcast programs featuring Person-Y. \newline
\textbf{Paraphrase}: Find Channel-X shows featuring Person-Y with 24 episodes. \newline
\textbf{Search-style}: Person-Y Channel-X 24-episode programs. \\[2pt]
Validation input &
Question outputs + gold Cypher + condition summary. \\
\bottomrule
\end{tabularx}
\caption{An anonymized data construction example before SFT conversion. The symbolic Cypher target is fixed before LLM rewriting, and validation checks the generated questions against the same target.}
\label{tab:data-construction-example}
\end{table*}
\section{Example SFT Instance}
\label{sec:sft-example}

Table~\ref{tab:sft-example} shows the candidate-aware SFT format used by the generator. The example illustrates the three pieces that are present in both training and serving: a natural-language question, relation and entity candidates, and the gold Cypher target. Entity identifiers are anonymized placeholders rather than real enterprise URIs.

\begin{table*}[p]
\centering
\small
\setlength{\tabcolsep}{4pt}
\renewcommand{\arraystretch}{1.15}
\begin{tabularx}{0.98\linewidth}{@{}p{0.14\linewidth}X@{}}
\toprule
Field & Example \\
\midrule

Instruction &
Convert the natural-language question into a Cypher query using the given candidate relations and entities. Output only the Cypher query. \\

\addlinespace[2pt]
Input &
\begin{minipage}[t]{\linewidth}
\raggedright
\textbf{Question}: Find 24-episode broadcast programs aired by Channel-X and featuring Person-Y.\\[2pt]

\textbf{Candidate relations}:\\
\ttfamily\small
{}[D] BROADCAST\_PROGRAM [N] cast\_member [R] PERSON [HINT] cast, actor\\
{}[D] BROADCAST\_PROGRAM [N] broadcast\_by [R] BROADCAST\_CHANNEL [HINT] broadcaster, channel\\
{}[D] BROADCAST\_PROGRAM [N] number\_of\_episodes [R] LITERAL [HINT] episode count\\
{}... other broadcast-program relations ...\\[2pt]
\normalfont
\textbf{Candidate entities}:\\
\ttfamily\small
{}[ID] <PERSON\_URI\_001> [N] Person-Y [C] PERSON\\
{}[ID] <CHANNEL\_URI\_001> [N] Channel-X [C] BROADCAST\_CHANNEL\\
{}... distractor entity candidates ...
\end{minipage} \\

\addlinespace[2pt]
Output &
\begin{minipage}[t]{\linewidth}
\raggedright\ttfamily\small
MATCH (n:BROADCAST\_PROGRAM)-[r0:cast\_member]->(m0:PERSON),\\
\phantom{MATCH }(n)-[r1:broadcast\_by]->(m1:BROADCAST\_CHANNEL),\\
\phantom{MATCH }(n)-[r2:number\_of\_episodes]->(m2:LITERAL)\\
WHERE m0.id = '<PERSON\_URI\_001>'\\
\phantom{WHERE }AND m1.id = '<CHANNEL\_URI\_001>'\\
\phantom{WHERE }AND m2.value\_number = 24\\
RETURN n, r0, m0, r1, m1, r2, m2
\end{minipage} \\
\bottomrule
\end{tabularx}
\caption{An anonymized SFT instance in the candidate-aware instruction format. The input contains the question, candidate relations, and entity candidates, and the output is the gold Cypher.}
\label{tab:sft-example}
\end{table*}

\end{document}